\documentclass[twoside,11pt]{article}

\usepackage{jmlr2e}
\usepackage{ifthen}
\usepackage{alltt}
\usepackage{color}
\usepackage{amsmath}
\usepackage{graphicx}
\usepackage{subfigure}
\usepackage[dvipsnames]{xcolor}
\hypersetup{
    colorlinks=true,
    citecolor=Violet,
    linkcolor=violet,
    filecolor=magenta,      
    urlcolor=BlueViolet
}
\colorlet{shadecolor}{gray!25}
\usepackage{listings}
\usepackage{multirow}
\providecommand{\doarxiv}{true}
\newboolean{isarxiv}
\setboolean{isarxiv}{\doarxiv} 
\ifthenelse{\boolean{isarxiv}}{
\newcommand{\arxiv}[1]{#1}
\newcommand{\notarxiv}[1]{}
}{
\newcommand{\arxiv}[1]{}
\newcommand{\notarxiv}[1]{#1}
}

\ShortHeadings{Jensen: An Easily-Extensible Toolkit for Convex
  Optimization and Machine Learning}{Iyer, Halloran, and Wei}
\firstpageno{1}

\begin{document}
\lstset{language=bash}
\title{Jensen: An Easily-Extensible C++ Toolkit for Production-Level
  Machine Learning and Convex Optimization}
% \title{Jensen: An Easily-Extensible C++ Toolkit for Large-Scale
%   Machine Learning and Convex Optimization}
% \title{CMLTK: A Flexible Toolkit for Efficient Convex Optimization and
%   Machine Learning with Extensive API Support}

\author{\name Rishabh Iyer$^1$ \email rishi@microsoft.com \\
       \name John T.\ Halloran$^2$ \email jthalloran@ucdavis.edu \\
       \name Kai Wei$^1$ \email kawe@microsoft.com \\
       \AND
       \addr $^1$Microsoft Corporation\\
       One Microsoft Way\\
       Redmond, WA 98052, USA
       \AND
       \addr $^2$Department of Public Health Sciences\\
       University of California, Davis\\
       Davis, CA 95616, USA
     }
% \author{\name Rishabh Iyer \email rishi@microsoft.com \\
%        \addr Microsoft Corporation\\
%        One Microsoft Way\\
%        Redmond, WA 98052, USA
%        \AND
%        \name John T.\ Halloran \email jthalloran@ucdavis.edu \\
%        \addr Department of Public Health Sciences\\
%        University of California, Davis\\
%        Davis, CA 95616, USA
%        \AND
%        \name Kai Wei \email kaiwei@uw.edu \\
%        \addr Microsoft Corporation\\
%        One Microsoft Way\\
%        Redmond, WA 98052, USA
%      }

\editor{Fill in later}

\maketitle

\begin{abstract}%   <- trailing '%' for backward compatibility of .sty file
This paper introduces Jensen, an easily extensible and scalable
toolkit for production-level machine learning and convex
optimization. Jensen implements a framework of convex (or loss)
functions, convex optimization algorithms (including Gradient
Descent, L-BFGS, Stochastic Gradient Descent, Conjugate Gradient,
etc.), and a family of machine learning classifiers and regressors
(Logistic Regression, SVMs, Least Square Regression, etc.). This
framework makes it possible to deploy and train models with a few
lines of code, and also extend and build upon this by integrating new
loss functions and optimization algorithms. 
\end{abstract}


\section{Introduction}

In the past few decades, convex optimization has emerged as a key
ingredient for machine learning.  With its growth and prominence,
several convex solvers have been released.  Such solvers broadly fall
into one of two categories: general solvers
% , for which very general convex
% objectives may be specified but are typically slow and limited to
% small-scale problems, 
and production-level solvers.  
% , which may quickly
% solve large-scale problems by narrowing the class of functions which
% may be represented.  
The former, such as CVX~\citep{cvx, cvxPy_2016}, are often written for
a high-level programming language (such as Matlab or Python) and may
be used to solve small-scale problems.  Such solvers are vital in instructional environments as
tools for developing practitioners to gain insight and intuition about
convex problems.  However, even modest-scale problems pose both
memory and runtime difficulties, while large-scale problems are
infeasible.  On the other hand, production-level software solve convex
problems for large-scale machine learning by narrowing the scope of
functions which may be represented.  For instance,
LIBSVM~\citep{libsvm_2011} offers non-linear support vector machine
(SVM) learning and, narrowing the scope of represented functions even
further, LIBLINEAR~\citep{liblinear_2008} offers faster model training
for the class of linear SVMs.
%  and, narrowing the scope of
% supported functions even further, LIBLINEAR~\citep{liblinear_2008}
% offers fast evaluation of linear support vector machines.

% In many cases, 
When the time comes for researchers to transition from
a general solver to developing their own domain-specific software, it
is highly desirable to call an ideally-suited production-level solver
using an API.  However, researchers are often met with a
number of obstacles when pursuing this route.  Firstly, owing to their
narrow-scope, a production-level solver may not exist for a general
objective initially investigated at small-scale using a general
solver.  This necessarily means developing a custom-solver from
scratch% , which is often the path pursued
.  Secondly, API support
varies widely and, in general, requires significant development time
and effort to merge into an existing project's framework% ; very often,
% user-friendly APIs do not exist and researchers are left with the
% daunting task of deciphering a large, external codebase
.  In order to
bridge the gap between general solvers and production-level solvers
while supporting an extremely user-friendly API, we present
Jensen. %, a toolkit for convex optimization and machine learning.

% However, API support widely
% varies and, in general, requires significant development time and
% effort to merge into an existing project's framework (user-friendly
% APIs often do not even exist, to begin with).  Furthermore,
% owing to the narrow scope of production-level software, many such
% transitions necessarily mean developing a custom-solver from scratch,
% as no production-level solver may exist for an objective initially
% investigated at small-scale using a general solver (this route is very
% often pursued rather than the large amount of effort required to
% decode a large, external codebase).  In order to bridge the gap between
% general solvers and production-level solvers, while providing a
% easy-to-use APIs, we present the Convex optimization and
% Machine Learning ToolKit (CMLTK).

Written in {\tt C++},  Jensen is an open-source, production-level
solver for general machine learning.  Users may define convex
functions, select from a large number of widely-used
optimization algorithms, and specify machine learning tasks to build
applications capable of handling problems at massive scale.  Supporting
general convex functions without sacrificing support of large-scale
problems, Jensen thus combines the strengths of production-level
software with those of general solvers.  

\section{Jensen}
Jensen allows the intuitive implementation of
new optimization algorithms using a blend of linear algebra (similar
to Matlab syntax) and {\tt C++} object-oriented syntax, similar to
(but not as specialized as) the domain-specific language (DSL) offered
by CVX.  We call this blend a pseudo-domain specific language (PDSL).
Coupled with the ability to construct and easily select among many
different optimization algorithms during runtime, Jensen adds a broad
degree of algorithmic design flexibility to suit different
application needs  (i.e., stochastic or batch learning, first order or
quasi-Newton, etc.) without rigorous recoding (demonstrated in
Section~\ref{section:softwareDesign}).

\subsection{Software Design}\label{section:softwareDesign}
The toolkit follows a highly modular design where different
optimization algorithms, objective functions, and machine learning
tasks may be combined to quickly produce tailor-made applications.
Training and testing (including cross-validation) for classification
(binary and multi-class) or regression are supported for a large
number of popular loss functions\arxiv{ (listed in Appendix Table~\ref{table:contFuncs})} and state-of-the-art optimization
algorithms\arxiv{ (listed in Appendix Table~\ref{table:optAlgs})}.
Algorithms may be intuitively and compactly expressed using Jensen's
PDSL, without any sacrifice in runtime performance.  For example,
gradient descent for arbitrary convex functions is defined in only 14
lines (illustrated below) and conjugate gradient descent for arbitrary
convex functions is defined in only 41 lines.

{\footnotesize
% \begin{lstlisting}[language=C++]
\begin{verbatim}
Vector gd(const ContinuousFunctions& c, const Vector& x0, const double alpha,
const int maxEval, const double TOL, const int verbosity){
        Vector x(x0), g;
        double f; // function value
        c.eval(x, f, g); // evaluate f(x), compute gradient g
        double gnorm = norm(g);
        int funcEval = 1;
        while ((gnorm >= TOL) && (funcEval < maxEval) ){
                multiplyAccumulate(x, alpha, g); // x = x - alpha*g
                c.eval(x, f, g);
                funcEval++;
                gnorm = norm(g);
        }
        return x;
}
\end{verbatim}
}
When building deployable machine learning applications, at the highest
design-level, machine learning modules call user-specified loss
functions.  During training, the loss function calls the desired
optimization algorithm.  For instance, using the OWL-QN
algorithm~\citep{andrew2007scalable}, the following trains then tests
an $l_1$-regularized, logistic regression classifier with training
file \texttt{trainFile} and testing file \texttt{testFile} in LIBSVM format:
{\footnotesize
\begin{verbatim}
int algtype = 0; // choose OWL-QN algorithm
double lambda = 1.0; // regularization strength
int ntrain, mtrain, ntest, mtest; // num training instances/features, test instances/features
vector<struct SparseFeature> trainFeatures, testFeatures; // feature vectors
Vector ytrain, ytest; // labels
readFeatureLabelsLibSVM(trainFile, trainFeatures, ytrain, ntrain, mtrain);
readFeatureLabelsLibSVM(testFile, testFeatures, ytest, ntest, mtest);
Classifiers<SparseFeature>* c = new
    L1LogisticRegression<SparseFeature>(trainFeatures, ytrain, mtrain, ntrain, nClasses, 
                                        lambda, algtype);
c->train();
accuracy = predictAccuracy(c, testFeatures, ytest);
\end{verbatim}
}

\noindent The solver-selection process is streamlined at the machine-learning
design level, where varying \texttt{algtype} in the earlier example
selects among various solvers.  The solvers themselves operate over
arbitrary continuous functions.  For instance, the earlier trained
classifier is equivalent to instantiating an $l_1$-regularized
logistic-loss function and passing this function to the desired
solver, as follows:
{\footnotesize
\begin{verbatim}
Vector x0 = Vector(mtrain, 0); // initial starting point
double alpha = 1.0; // initial step size
double gamma = 1e-4; // back-tracking parameter used by line-search solvers
int maxIter = 250;
int lbfgsMemory = 100; // budget for limited-memory solvers
double eps = 1e-3; // convergence criteria
L1LogisticLoss<Feature> ll(mtrain, trainFeatures, ytrain, lambda);
lbfgsMinOwl(ll, x0, alpha, gamma, maxIter, lbfgsMemory, eps);
\end{verbatim}
}

Each module is self-contained and may be used for general tasks; at
its core, Jensen is a solver library optimizing arbitrary continuous
functions using general optimization algorithms.  Thus, Jensen may not
only be used to quickly deploy new machine learning applications, but
also to easily investigate convex optimization problems in an
instructional setting.  For instance, the earlier call to
\texttt{lbfgsMinOwl} may be replaced with an out-of-the-box solver or
replaced with a user-constructed solver written using Jensen's PDSL.
Note the ease with which the APIs of Jensen data types are called
in the previous examples.\arxiv{  The major abstract data types are further
detailed in Appendix Section~\ref{section:dataTypes}.}

\section{Experiments}\label{section:experiments}
% We compare the speed of CMLTK to the general solver CVXPY and production-level
% solvers LibSVM and LIBLINEAR.  Each solver is used to optimize several
% popular machine learning objective functions using three datasets ranging from
% small to large-scale: the small-scale dataset ijcnn1 containing 49,990 data
% instances and 22 features, the large-scale dataset rcv1\_train
% containing 20,242 data instances and 677,399 features, and the
% large-scale dataset news20 containing 19,996 data instances and
% 1,355,191 features.  Each
% dataset is freely available for download at the
% \href{https://www.csie.ntu.edu.tw/~cjlin/libsvmtools/datasets/binary.html}{LibSVM
%  Data page}.  
We compare the speed of Jensen to the general solver CVXPY and production-level
solvers LIBLINEAR and LIBSVM using three datasets ranging from
small- to large-scale: the small-scale dataset ijcnn1 containing
22 features and 35,000 training instances, the mid-scale dataset rcv1\_train
containing 677,399 features and 20,242 training instances, and the
large-scale dataset news20 containing 1,355,191 features and 19,996
training instances.  Each
dataset is freely available for download at the
\href{https://www.csie.ntu.edu.tw/~cjlin/libsvmtools/datasets/binary.html}{LibSVM
 Data page}.  

\begin{table}[h]
\centering
{\footnotesize
\begin{tabular}{|c||c|c|c|c|c|}
\hline
Dataset & Objective & Jensen & LIBLINEAR & CVXPY & LIBSVM\\\hline
\multirow{5}{*}{ijcnn1}
% maxiter 50, eps 0.01
% LR = TRON, SVM Primal = L-BFGS, SVM Dual = SVCDual
& Logistic Regression & 0.50 & 0.27 & 221.72 & *\\\cline{2-6}
& L2-SVM, Primal & 0.64 & 0.23 & 4.03 & *\\\cline{2-6}
& L2-SVM, Dual & 0.51 & 0.39 & $\star$ & *\\\cline{2-6}
& L1-SVM, Primal & 0.45 & * & 4.13 & * \\\cline{2-6}
& L1-SVM, Dual & 0.50 & 0.31 & 6.04 & 14.74\\\hline
\multirow{4}{*}{rcv1\_train} 
& Logistic Regression & 1.27 & 0.85 & 270.16 & *\\\cline{2-6}
& L2-SVM, Primal & 2.26 & 0.66 & 31.83 & *\\\cline{2-6}
& L2-SVM, Dual & 0.97 & 0.77 & $\dagger$ & *\\\cline{2-6}
& L1-SVM, Primal & 1.88 & * & 31.92 & * \\\cline{2-6}
& L1-SVM, Dual & 0.95 & 1.11 & $\dagger$ & 52.78\\\hline
\multirow{4}{*}{news20} 
& Logistic Regression & 7.64 & 4.80 & $\star$ & *\\\cline{2-6}
& L2-SVM, Primal & 28.69 & 5.81 & $\star$ & *\\\cline{2-6}
& L2-SVM, Dual & 4.72 & 4.09 & $\dagger$ & *\\\cline{2-6}
& L1-SVM, Primal & 23.40 & * & $\dagger$ & * \\\cline{2-6}
& L1-SVM, Dual & 4.71 & 5.55 & $\dagger$ & 318.29\\\hline
\end{tabular}
\caption{{\footnotesize Solver runtimes (in seconds) for several popular training
  objective functions evaluated over datasets of increasing scale.
  As used by LIBLINEAR, L1-SVM denotes the SVM using hinge-loss and
  L2-SVM denotes the SVM using the squared hinge-loss.  Each
  formulation is $l_2$-regularized.  * denotes
formulations not supported by the solver, $\dagger$ denotes
formulations for CVXPY exhausted all system memory and were infeasible, and
$\star$ denotes formulations for which CVXPY exited
before completion without further diagnostic detail.}
% which were infeasible due to CVXPY exhausting all system memory
%  (in the case
% of solving a primal L2-SVM, this occurred after ten days of continuous
% computation)
}
}
\label{table:runtimes}
\end{table}

Runtimes are reported in Table~\ref{table:runtimes} evaluating
several popular machine learning objective functions, where each
objective includes $l_2$-regularization.
All tests were run on the same machine with an
Intel Xeon E5-2620 (clocked at 2.1 GHz) and 64 GB of memory.
Tests were run using LIBLINEAR version 2.11, LIBSVM version 3.22, and
CVXPY version 0.4.10.  Jensen,
LIBSVM, and LIBLINEAR wall-clock runtimes are reported.  Due to the
excessive overhead incurred by formulating problems
in CVXPY (e.g., the dual L1-SVM objective for small-scale dataset
ijcnnl required 3.46 hours to formulate), CVXPY runtimes report only the
elapsed runtime calling a formulated problem's \texttt{solve()}
attribute.  Jensen experiments were run using several different
optimization algorithms to demonstrate the toolkit's breadth:
L-BFGS~\citep{wright1999numerical} for primal SVM objectives,
TRON~\citep{lin2008trust} for logistic regression, and dual coordinate
descent~\citep{hsieh2008dual} for dual SVM objectives.  All method's
were run with $\epsilon=0.01$; however, dual problems in CVXPY were
subject to numerical issues, requiring $\epsilon=1e-5$ to obtain
accurate solutions.  Logistic regression and primal SVM experiments
using CVXPY were run using solver \texttt{SCS}, which was found to be
faster than other CVXPY solvers (three orders-of-magnitude faster in
some cases), while dual SVM experiments were run using solver
\texttt{ECOS} (found to be slightly faster than \texttt{SCS} for this
problem).  The training accuracy for each completed method's learned
parameters performed similarly, agreeing up to two significant figures for all
completed methods and respective datasets with the exception of CVPY
achieving a one percent higher accuracy than competitors for logistic regression on dataset rcv1.

\section{Conclusions}
We've described Jensen, a flexible open-source library for scalable
machine learning and convex optimization. The source code of Jensen is available at \url{https://github.com/rishabhk108/jensen}. Jensen does not rely on external packages and is supported on Unix, Windows, and OSX Operating Systems.

\section*{Acknowledgments}
This work was partially supported by NIH NCATS grant UL1 TR001860.

\bibliography{main}
\pagebreak
\begin{appendix}

\section{Jensen Loss Functions and Optimization Algorithms}
\begin{table}[h]
\begin{center}
{\small
\begin{tabular}{|c||c|}
\hline
Loss function & Description\\\hline
\texttt{L2LeastSquaresLoss} & $l_2$-regularized, least squares loss.\\\hline
\texttt{L1LeastSquaresLoss} & $l_1$-regularized, least squares loss.\\\hline
\texttt{L2LogisticLoss} & $l_2$-regularized, logistic loss.\\\hline
\texttt{L1LogisticLoss} & $l_1$-regularized, logistic loss.\\\hline
\texttt{L2ProbitLoss} & $l_2$-regularized, probit loss.\\\hline
\texttt{L1ProbitLoss} & $l_1$-regularized, probit loss.\\\hline
\texttt{L2HingeSVMLoss} & $l_2$-regularized, linear L1-SVM
  loss (i.e., hinge loss).\\\hline
\texttt{L1HingeSVMLoss} & $l_1$-regularized, linear L1-SVM
loss.\\\hline
\texttt{L2SmoothSVMLoss} & $l_2$-regularized, linear L2-support vector
machine loss (i.e., squared
hinge loss).\\\hline
\texttt{L1SmoothSVMLoss} & $l_1$-regularized, linear L2-SVM
loss.\\\hline
\texttt{L2HuberSVMLoss} & $l_2$-regularized, linear SVM
  with Huber loss.\\\hline
\texttt{L1HuberSVMLoss} & $l_1$-regularized, linear SVM
with Huber loss.\\\hline
\texttt{L2HingeSVRLoss} & $l_2$-regularized, linear L1-support vector
regression (SVR) loss.\\\hline
\texttt{L1HingeSVRLoss} & $l_1$-regularized, linear L1-SVR
loss.\\\hline
\texttt{L2SmoothSVRLoss} & $l_2$-regularized, linear L2-SVR loss.\\\hline
\texttt{L1SmoothSVRLoss} & $l_1$-regularized, linear L2-SVR loss.\\\hline
\end{tabular}
}
\caption{Out-of-the-box Jensen loss functions.  L1- and L2-SVMs are
  described at length in \cite{liblinear_2008}, where L1-SVM utilizes
  the standard hinge loss and L2-SVM uses the hinge loss squared
  (rendering it differentiable).}
\label{table:contFuncs}
\end{center}
\end{table}

\begin{table}[h]
\begin{center}
{\footnotesize
% \begin{tabular}{|p{2.3in}||p{3.3in}|}
\begin{tabular}{|p{2.3in}||p{3.5in}|}
\hline
Algorithm name & Description\\\hline
\texttt{gd} & Gradient descent (GD) with static step size.\\\hline
\texttt{gdLineSearch} & GD with backtracking line
search.\\\hline%~\citep{armijo1966minimization}.\\\hline
\texttt{gdBarzilaiBorwein} & GD with backtracking
line search and Barzilai-Borwein step
length.\\\hline%~\citep{barzilai1988two}.\\\hline
\texttt{gdNesterov} & GD with Nesterov's accelerated
method.\\\hline%~\citep{nesterov1983method}.\\\hline
\texttt{lbfgsMin} & L-BFGS for Quasi-Newton optimization~\citep{wright1999numerical}.
\\\hline
\texttt{lbfgsMinOwl} & Orthant-Wise Limited-memory
Quasi-Newton(OWL-QN)~\citep{andrew2007scalable} for $l_1$ regularized
Quasi-Newton optimization.
\\\hline
\texttt{tron} & Trust Region Newton~\citep{lin2008trust}.
\\\hline
\texttt{SVCDual} & Dual coordinate descent algorithm for SVMs~\citep{hsieh2008dual}.
\\\hline
% \texttt{L1LRPrimal} & Primal coordinate descent algorithm for $l_1$
% regularized logistic regression~\citep{yuan2010comparison}.\\\hline
\texttt{sgd} & Stochastic gradient descent (SGD) with static step size.\\\hline
\texttt{sgdDecayingLearningRate} & SGD with a
specified rate of decay.\\\hline
\texttt{sgdAdagrad} & SGD with AdaGrad~\citep{duchi2011adaptive}\\\hline
\texttt{sgdStochasticAverageGradient} & Stochastic average gradient
(SAG) descent~\citep{roux2012stochastic}.\\\hline
\texttt{sgdRegularizedDualAveraging} & SGD using the regularized
dual-averaging algorithm~\citep{xiao2010dual} for $l_1$
regularization.\\\hline
\texttt{sgdRegularizedDualAveragingAdagrad} & SGD using the regularized
dual-averaging algorithm and AdaGard.\\\hline
\end{tabular}
}
\caption{Out-of-the-box Jensen optimization algorithms.}
\label{table:optAlgs}
\end{center}
\end{table}

\section{Jensen Data Types}\label{section:dataTypes}
\subsection{Objective (Loss)  Functions}
The continuous function abstract base class defines operations such as
function evaluation, computation of the (sub) gradient, and
computation of the (sub) Hessian.  Derived
classes correspond to loss (or arbitrary objective) functions 
such as $l_1$-regularized logistic loss, $l_2$-regularized hinge
loss, and so on.  Once a new loss function is derived, any Jensen
optimization algorithm (discussed in
Section~\ref{section:optAlgorithms}) may be used. Natively implemented
loss functions are listed in Table~\ref{table:contFuncs}.

\subsection{Optimization Algorithms}\label{section:optAlgorithms}
Optimization algorithms are defined to act on continuous functions.
This generality means we need only define
a specific algorithm once to optimize any continuous loss functions.
This design substantially cuts down debugging and coding time while
greatly streamlining the codebase (the optimization algorithms
themselves are also very compact).  This also means that once a user
instantiates a loss function, any Jensen optimization algorithm may be
used as a solver.  Natively implemented optimization algorithms are
listed in Table~\ref{table:optAlgs}.

\subsection{Machine Learning}
With the efficient optimization of arbitrary loss functions, machine
learning is readily applicable for real-world problems.  In Jensen,
(multi-class) classification and regression are natively supported so
that classifiers and regressors may be customized as
desired.  For instance, if classification using $l_1$-regularized
logistic regression trained using the OWL-QN algorithm is desired, all
that is necessary is to instantiate continuous
function \texttt{L1LogisticLoss} and specify the \texttt{lbfgsMinOwl}
solver; this example may be found in\\
\texttt{\$JENSEN/src/machinelearning/logisticRegression/L1LogisticRegression.cc},
where \texttt{\$JENSEN} is the uncompressed download directory
(further discussed in Appendix Section~\ref{appendix:building}).

\section{Build Process and Examples}\label{appendix:building}
For the discussion that follows, we assume that the toolkit
source has been cloned (or uncompressed) to directory
\texttt{\$JENSEN} and that commands are run using Bash (denoted by a
preceding \texttt{\$}).  Linking and
building for Jensen are handled using \texttt{CMake}.  After
downloading the codebase, the following builds the example
applications found in \texttt{\$JENSEN/test} to directory
\texttt{\$JENSEN/build}:
\begin{lstlisting}[language=bash, backgroundcolor =
  \color{shadecolor}]
$ cd $JENSEN
$ mkdir build
$ cd build
$ cmake .. && make
\end{lstlisting}

The examples in \texttt{\$JENSEN/test} demonstrate the API use for all
out-of-the-box loss functions (listed in Table~\ref{table:contFuncs}) and
optimization algorithms (listed in Table~\ref{table:optAlgs})
evaluated over real-world data (in \texttt{\$JENSEN/data}).  For
instance, \texttt{\$JENSEN/test/TestL2LogisticLoss.cc} displays how to
instantiate the loss function for $l_2$-regularized logistic
regression and optimize this function using all relevant
out-of-the-box solvers (i.e., gradient descent, TRON, L-BFGS,
stochastic gradient descent, etc.).  From
\texttt{\$JENSEN/test/TestL2LogisticLoss.cc}, the
following code excerpt loads the dataset, instantiates the loss
function, and uses gradient descent to optimize the loss function:
{\small
\begin{alltt}
\vdots
char* featureFile = "../data/20newsgroup.feat";
char* labelFile = "../data/20newsgroup.label";
int n; // number of data items
int m; // numFeatures
vector<struct SparseFeature> features = readFeatureVectorSparse(featureFile, n, m);
Vector y = readVector(labelFile, n);
L2LogisticLoss<SparseFeature> ll(m, features, y, 1);

int numEpochs = 50;
double stepSize = 1e-5;
double f;
Vector x0(m, 0), g, x;
\vdots
x = gd(ll, x0, stepSize, numEpochs);
\end{alltt}}
\noindent After linking and building, running the binary
\texttt{\$JENSEN/build/TestL2LogisticLoss} will proceed optimizing the
$l_2$-regularized logistic loss function with a
large number of optimization algorithms.  Similar respective binaries
are built for all loss functions (and applicable out-of-the-box
optimization algorithms) listed in Table~\ref{table:contFuncs}.

The use of \texttt{CMake} makes linking and building user-customized
modules simple; in the previous example, linking the $l_2$-regularized
logistic loss function and defining the test executable to be built
are specified in \texttt{\$JENSEN/CMakeLists.txt}:
{\small
\begin{alltt}
add_library(jensen
\vdots
src/optimization/contFunctions/L2LogisticLoss.cc
\vdots
)

add_executable(TestL2LogisticLoss test/TestL2LogisticLoss.cc)
target_link_libraries(TestL2LogisticLoss jensen)
\end{alltt}
}
\noindent New user-defined loss functions, optimization algorithms, and machine
learning applications may be easily added, linked to the toolkit library, and built in
a similar manner.

\subsection{Machine Learning Application Examples}
Examples of building deployable applications using Jensen's
machine learning APIs (which combine arbitrary loss functions and
optimization algorithms to perform machine learning tasks) and
command line/file-handling utilities are also available in
\texttt{\$JENSEN/test}.  For example,
\texttt{\$JENSEN/test/TestClassification.cc} builds a classifier which
accepts as command-line input training files, testing files, the loss function to
be evaluated, the optimization algorithm to be utilized, and various
optimization algorithm options.  After linking and building, running
the following will train then evaluate the accuracy of an
$l_1$-regularized L2-SVM classifier (with regularization strength
$\lambda = 0.25$) trained using the OWL-QN
algorithm~\citep{andrew2007scalable}:
\begin{lstlisting}[language=bash, backgroundcolor =
  \color{shadecolor}]
$ cd $JENSEN/build
$ ./TestClassification \
-method 3 -algtype 0 -reg 0.25 -nClasses 20 \
-maxIter 1000 -startwith1 true \
-trainFeatureFile ../data/20newsgroup.feat \
-trainLabelFile ../data/20newsgroup.label \
-testFeatureFile ../data/20newsgroup.feat \
-testLabelFile ../data/20newsgroup.label
\end{lstlisting}

In the above example, varying \texttt{method} alters the objective
function used for training and testing and varying \texttt{algtype}
alters the training algorithm utilized.  Thus, the generality and
flexibility of the toolkit facilitate the rapid evaluation of different
objective functions and training algorithms given new data.  Note that the
source file for this example is easily customizable (including
command-line options) and serves as a template for users to easily
create tailor-made applications for large-scale, real-world data.
Other deployable application examples in \texttt{\$JENSEN/test}
include \texttt{TestClassificationCrossVal.cc} and
\texttt{TestClassificationLibSVM.cc}, which demonstrate
cross-validation and LIBSVM file-format support, respectively.

\end{appendix}

\end{document}